\documentclass[conference]{IEEEtran}
\IEEEoverridecommandlockouts
\usepackage{cite}
\usepackage{url}
\usepackage{amsmath,amssymb,amsfonts}
\usepackage{dsfont}
\usepackage{graphicx}
\usepackage{textcomp}
\usepackage{xcolor}
\def\BibTeX{{\rm B\kern-.05em{\sc i\kern-.025em b}\kern-.08em
    T\kern-.1667em\lower.7ex\hbox{E}\kern-.125emX}}
    
\usepackage{tikz}
\usepackage{pgfplots}
\usetikzlibrary{arrows}
\usetikzlibrary{shapes}
\usetikzlibrary{shapes.misc}
\usetikzlibrary{tikzmark}
\usetikzlibrary{matrix, positioning}
\usetikzlibrary{backgrounds, intersections, decorations.pathreplacing}

\tikzstyle{edge}=[->, >=stealth', shorten <=2pt, shorten >=2pt, auto, line width=0.5mm]
    
\usepackage{algorithm}
\usepackage{algorithmicx}
\usepackage{algpseudocode}
    
\usepackage{multirow}
\usepackage{multicol}
\usepackage{placeins}
    
\newcommand{\refdef}[1]{Definition~\ref{#1}}    
\newcommand{\refeq}[1]{Eq.~\eqref{#1}}

\newcommand{\dimsym}{d}
\newcommand{\N}{\mathbb{N}}
\newcommand{\RN}{\mathbb{R}}
\DeclareMathOperator*{\argmin}{\arg\min}    
\DeclareMathOperator*{\loss}{{\ell}}

\newcommand{\mat}[1]{\mathbf{#1}}
\newcommand{\set}[1]{\mathcal{#1}}
\newcommand{\pnorm}[1]{\lVert{#1}\rVert}    
\DeclareMathOperator*{\regularization}{\ensuremath{{\theta}}}
\DeclareMathOperator*{\diversity}{\ensuremath{{\psi}}}
\newcommand{\classifier}{\ensuremath{h}}
\newcommand{\x}{\ensuremath{\vec{x}}}
\newcommand{\xorig}{\ensuremath{\vec{x}_\text{orig}}}
\newcommand{\y}{\ensuremath{\vec{y}}}
\newcommand{\z}{\ensuremath{\vec{z}}}  
\newcommand{\xcf}{\ensuremath{\vec{x}_{\text{cf}}}}   
\newcommand{\deltacf}{\ensuremath{\vec{\delta}_{\text{cf}}}}    
\newcommand{\ycf}{\ensuremath{\vec{y}_{\text{cf}}}}
\newcommand{\ycfclassifier}{\ensuremath{y_{\text{cf}}}}
\newcommand{\setX}{\ensuremath{\set{X}}}  
\newcommand{\setY}{\ensuremath{\set{Y}}}  
\newcommand{\prototype}{\vec{p}}
  
\newcommand{\setI}{\ensuremath{\set{I}}} 
\newcommand{\setF}{\ensuremath{\set{F}}} 
\newcommand{\encoder}{\text{enc}}
\newcommand{\decoder}{\text{dec}}
\newcommand{\params}{\ensuremath{\theta}}
\newcommand{\mlp}{\ensuremath{f}}
\newcommand{\myCF}[4]{\ensuremath{\text{CF}_{#4}(#1,#2, #3)}}

\newcommand{\dimred}{\ensuremath{\phi}}
\newcommand{\DR}{DR}
\newcommand{\smth}{something}
    
\newtheorem{definition}{Definition}
    
\begin{document}

\title{``Why Here and Not There?'' -- Diverse Contrasting Explanations of Dimensionality Reduction\\
\thanks{We gratefully acknowledge funding from the VW-Foundation for the project \textit{IMPACT} funded in the frame of the funding line \textit{AI and its Implications for Future Society}.

Andr\'e Artelt is also affiliated with KIOS Research and Innovation Center of Excellence, University of Cyprus, Nicosia, Cyprus.}
}

\author{\IEEEauthorblockN{1\textsuperscript{st} Andr\'e Artelt}
\IEEEauthorblockA{\textit{Faculty of Technology} \\
\textit{Bielefeld University}\\
Bielefeld, Germany \\
aartelt@techfak.uni-bielefeld.de}
\and
\IEEEauthorblockN{2\textsuperscript{nd} Alexander Schulz}
\IEEEauthorblockA{\textit{Faculty of Technology} \\
\textit{Bielefeld University}\\
Bielefeld, Germany \\
aschulz@techfak.uni-bielefeld.de}
\and
\IEEEauthorblockN{3\textsuperscript{rd} Barbara Hammer}
\IEEEauthorblockA{\textit{Faculty of Technology} \\
\textit{Bielefeld University}\\
Bielefeld, Germany \\
bhammer@techfak.uni-bielefeld.de}
}

\maketitle

\begin{abstract}
Dimensionality reduction is a popular preprocessing and a widely used tool in data mining. Transparency, which is usually achieved by means of explanations, is nowadays a widely accepted and crucial requirement of machine learning based systems like classifiers and recommender systems. However, transparency of dimensionality reduction and other data mining tools have not been considered in much depth yet, still it is crucial to understand their behavior -- in particular practitioners might want to understand why a specific sample got mapped to a specific location.

In order to (locally) understand the behavior of a given dimensionality reduction method, we introduce the abstract concept of contrasting explanations for dimensionality reduction, and apply a realization of this concept to the specific application of explaining two dimensional data visualization.
\end{abstract}

\begin{IEEEkeywords}
XAI, Dimensionality reduction, Data visualization, Counterfactual explanations, Data mining
\end{IEEEkeywords}

\section{Introduction}\label{sec:intro}
Transparency of machine learning (ML) based system, applied in the real world, is nowadays a widely accepted requirement -- the importance of transparency was also recognized by the policy makers and therefore made its way into legal regulations like the EU's GDPR~\cite{GDPR}. A popular way of achieving transparency is by means of explanations~\cite{molnar2019} which then gave rise to the field of eXplainable AI (XAI)~\cite{ExplainableArtificialIntelligence,SurveyXai}. Although a lot of different explanation methodologies for ML based systems have been developed~\cite{molnar2019,SurveyXai}, it is important to realize that it is still somewhat unclear what exactly makes up a good explanation~\cite{doshivelez2017rigorous,offert2017i}. Therefore one must carefully pick the right explanation in the right situation, as there are (potentially) different target users with different goals~\cite{ribera2019can} -- e.g. ML engineers need explanations that help them to improve the system, while lay users need trust building explanations.
Popular explanations methods~\cite{molnar2019,SurveyXai} are feature relevance/importance methods~\cite{FeatureImportance}, and examples based methods~\cite{CaseBasedReasoning} which use a set or a single example for explaining the behavior of the system. Instances of example based methods are contrasting explanations like counterfactual explanations~\cite{CounterfactualWachter,CounterfactualReviewChallenges} and prototypes \& criticisms~\cite{PrototypesCriticism}.

Dimensionality reduction methods are a popular tool in data mining -- e.g. data visualization -- and an often used preprocessing in general ML pipelines~\cite{gisbrecht2015data}. Similar to other ML methods, dimensionality reduction methods itself are not easy to understand -- i.e. a high-dimensional sample is ``somehow'' mapped to a low-dimensional sample without providing any explanation/reason of this mapping.
A ML pipeline can not be transparent if it contains non-transparent preprocessings like dimensionality reduction, and a proper and responsible use of data analysis tools such as data visualization is not possible if the inner working of the tool is not understood. Therefore, we argue that there is a need for understanding dimensionality reduction methods -- we aim to provide such an understanding by means of contrasting explanations. 

\paragraph*{Related work}
In the context of explaining dimensionality reduction, only little work exists so far. Some approaches \cite{schulz2015metric,schulz2014relevance} 
aim to infer global feature importance for a given data projection. Another work \cite{bibal2020explaining} 
estimates feature importance locally for a vicinity around a projected data point, using locally linear models. 
A recent paper~\cite{bardos2022local} proposes to use local feature importance explanations by computing a local linear approximation for each reduced dimension, extracting feature importances from the weight vectors.
Further, saliency map approaches such as the layer-wise relevance propagation (LRP) \cite{bach2015pixel} could in principle be applied to a parametric dimensionality reduction mapping in order to obtain locally relevant features. However, these approaches do not provide contrasting explanations, in which we are interested here.

\paragraph*{Our contributions}
First, we make a conceptional contribution by proposing a general formalization of diverse counterfactual explanations for explaining dimensionality reduction methods.
Second, we propose concrete realizations of this concept for four popular representatives of parametric dimensionality reduction method classes: PCA (linear mappings), SOM \cite{kohonen1990self} (topographic mappings), autoencoders \cite{goodfellow2016deep} (neural networks) and parametric t-SNE \cite{van2009learning} (parametric extensions of neighbor embedders). 
Finally, we empirically evaluate them in the particular use-case of two-dimensional data visualization.

\begin{figure*}
\begin{tikzpicture}[level/.style={thick},scale=1]
    \node (hd) at (0,0) {\includegraphics[scale=.26, trim=0 5cm 0cm 6cm, clip=true]{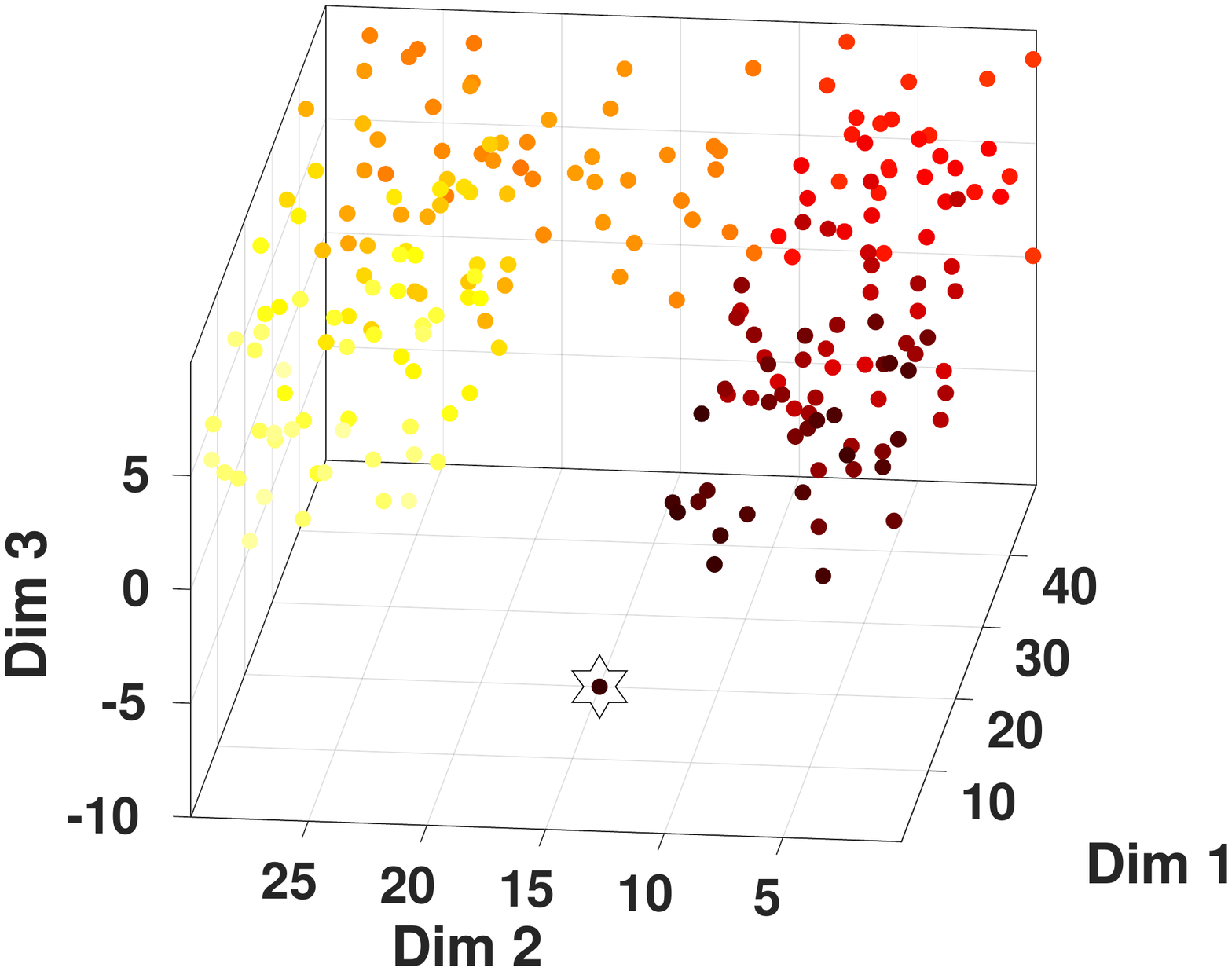}};
    \node (ld) at (6.5,0) {\includegraphics[scale=.26, trim=1cm 5cm 0 6cm, clip=true]{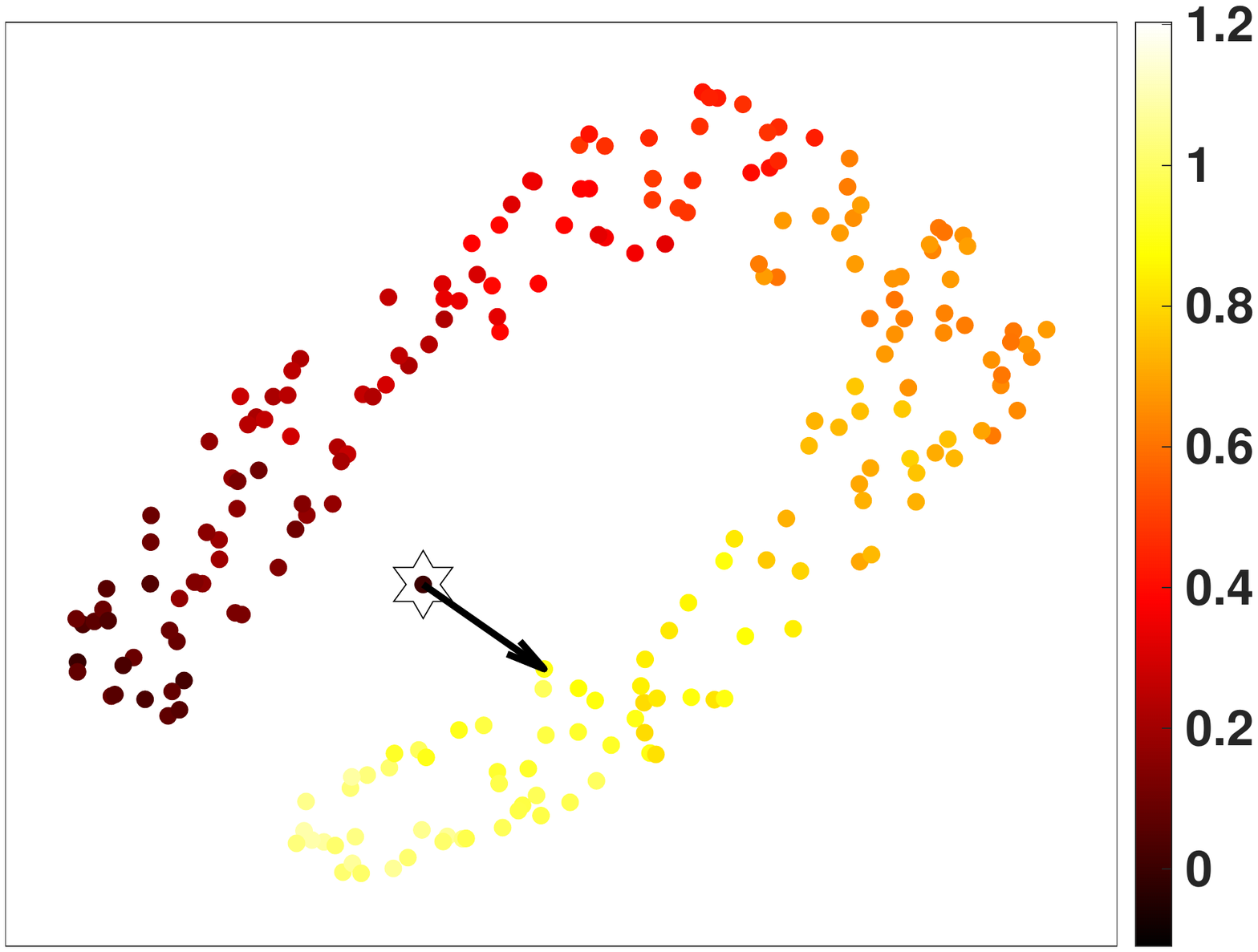}};
    \node[align=left] (expl1) at (12.7,0.75) {dim2 is not 20\\and dim3 is not 5};
    \node (expl_or) at (11.5,0) {or};
    \node[align=left] (expl2) at (12.7,-0.75) {dim2 is not 20\\and dim3 is not 0};
     
    \draw[edge,shorten <=-3pt, shorten >=-3pt] (hd.east) to[bend left]  node[above] {\DR} (ld.west);
    \draw[edge] (ld.east) to[bend left]  node[above] {Explanation} (expl_or.west);
    
\end{tikzpicture}
\caption{Illustration of the investigated topic: a 'high-dimensional' data set (left, with an outlier marked as a star) is mapped to two dimensions (middle), where the question 'why is the central point mapped here and not there' is asked (indicated by the arrow). Possible explanations are depicted (right).}
\label{fig:idea}
\end{figure*}

The remainder of this work is structured as follows: First (Section~\ref{sec:foundations}) we review the necessary foundations of dimensionality reduction and contrasting explanations. Next (Section~\ref{sec:modeling}), we propose and formalize diverse counterfactual explanations for explaining dimensionality reduction -- we first propose a general concept (Section~\ref{sec:modeling}), and then propose practical realizations for popular parametric dimensionality reduction methods (Sections~\ref{sec:modelspecific},\ref{sec:diverse_counterfactuals}). We empirically evaluate our proposed explanations in Section~\ref{sec:experiments} where we consider two-dimensional data visualization as a popular application of dimensionality reduction. Finally, this work closes with a summary and conclusion in Section~\ref{sec:conclusion}.

\section{Foundations}\label{sec:foundations}
\subsection{Dimensionality Reduction}\label{sec:foundations:dimred}
The common setting for dimensionality reduction (\DR) is that data $\x_i, i=1,\dots,m$ are given in a high-dimensional input space $\setX$ -- we will assume $\setX = \RN^\dimsym$ in the following. The goal is to project them to lower-dimensional points $\y_i,i=1,\dots,m$ in $\RN^{\dimsym'}$ -- where for data visualization often $\dimsym'=2$ --, such that as much structure as possible is preserved. The precise mathematical formalization of the term ``structure preservation'' is then one of the key differences between different \DR\ methods in literature \cite{van2009dimensionality,lee2007nonlinear,bunte2012general}.

One major view for grouping \DR\ methods is whether they provide an explicit function $\dimred: \setX \to \RN^{\dimsym'}$ for projection, where the parameters of $\dimred$ are adjusted by the according \DR\ method, or whether no such functional form is assumed by the approach. The former methods are referred to as parametric and the latter ones as non-parametric \cite{van2009dimensionality,gisbrecht2015data}.

Since we require parametric mappings in our work, we recap a few of the most popular parametric \DR\ approaches in the following. However, since there do exist successful extensions for non-parametric approaches to also provide a parametric function, we will consider one of them here as well.
We will consider these approaches again in our experiments.

\subsubsection{Linear Methods}\label{sec:linMap}
The most classical \DR\ methods are based on a linear functional form:
\begin{equation}\label{eq:dimred:linear}
    \dimred(\x) = \mat{A} \x + \vec{b}
\end{equation}
where $\mat{A}\in\RN^{\dimsym'\times\dimsym}$ and $\vec{b}\in\RN^{\dimsym'}$. Particular instances are Principal Component Analysis (PCA), Linear Discriminant Analysis (LDA) and also the mappings obtained by metric learning approaches such as the Large Margin Nearest Neighbor (LMNN) method \cite{gisbrecht2015data}. These constitute different cost function based approaches for estimating the parameters of $\dimred(\cdot)$, but in the end result in such a linear parametric mapping~\refeq{eq:dimred:linear}.

\subsubsection{Topographic Mappings}\label{sec:SOM}
A class of non-linear \DR\ approaches is given by topographic mappings such as the Self Organizing Map (SOM) and the Generative Topographic Mapping (GTM). We consider the SOM as one representative of this class of methods in the following. 
The SOM~\cite{kohonen1990self} consists of a set of prototypes $\prototype_{\z}\in\RN^\dimsym$ which are mapped to an index set $\setI$, $\dimred: \RN^\dimsym \to \setI$ -- e.g. the prototypes are arranged as a two-dimensional grid: $\setI \subset \N^2$.
The dimensionality reduction maps a given input $\x$ to the index of the closest prototype:
\begin{equation}\label{eq:dimred:som}
    \dimred(\x) = \underset{\z\,\in\,\setI}{\argmin}\; \pnorm{\x - \prototype_{\z}}_{2}.
\end{equation}

\subsubsection{Autoencoder}\label{sec:autoenc}
An autoencoder (AE) $\mlp_{\params}: \RN^\dimsym \to \RN^\dimsym$ is a neural network consisting of an encoder, mapping the input to a smaller representation (also called the bottleneck) and a decoder, mapping it back to the original input~\cite{goodfellow2016deep}:
\begin{equation}
    \mlp_{\params}(\x) = (\decoder_{\params} \circ \encoder_{\params})(\x),
\end{equation}
which are trained to optimize the reconstruction loss.
A (typically non-linear) dimensionality reduction $\dimred(\cdot)$ based on this approach consists of the encoder mapping:
\begin{equation}
    \dimred(\x) = \encoder_{\params}(\x)
\end{equation}

\subsubsection{Neighbor Embeddings}\label{sec:tsne}
The class of neighbor embedding methods constitutes a set of non-parametric approaches that are considered as the most successful or state-of-the-art techniques in many cases \cite{kobak2019art,becht2019dimensionality,gisbrecht2015data}. Instances are the very popular t-Distributed Stochastic Neighbor Embedding (t-SNE) and Uniform Manifold Approximation and Projection for Dimension Reduction (UMAP) approaches  \cite{van2008visualizing,mcinnes2018umap}.
In the following, we consider t-SNE as a representative of this class of methods and, among its parametric extensions \cite{van2009learning,gisbrecht2015parametric}, the approach Parametric t-SNE~\cite{van2009learning}.

Parametric t-SNE, uses a neural network $\mlp_{\params}: \RN^\dimsym \to \RN^{\dimsym'}$ for mapping a given input $\x$ to a lower-dimensional domain:
\begin{equation}
    \dimred(\x) = \mlp_{\params}(\x),
\end{equation}
with respect to the t-SNE cost function.

While there do exist more families of \DR\ approaches, such as manifold embeddings (including MVU and LLE) or discriminative/supervised \DR, it would exceed the scope of the present work to investigate all possible choices.

\subsection{Contrasting Explanations}\label{sec:foundations:cf}
Contrasting explanations state a change to some features of a given input such that the resulting data point causes a different behavior of the system/model than the original input does. Counterfactual explanations (often just called \textit{counterfactuals}) are the most prominent instance of contrasting explanations~\cite{molnar2019}. One can think of a counterfactual explanation as a recommendation of actions that change the model's behavior/prediction. One reason why counterfactual explanations are so popular is that there exists evidence that explanations used by humans are often contrasting in nature~\cite{CounterfactualsHumanReasoning} -- i.e. people often ask questions like \textit{``What would have to be different in order to observe a different outcome?''}. It was also shown that such questions are useful to learn about an unknown functionality and exploit this knowledge to achieve some goals~\cite{kuhl2022keep,kuhl2022let}.

The most prominent example from literature for illustrating the concept of a counterfactual explanation is the example of loan application: \textit{Imagine you applied for a loan at a bank. Unfortunately, the bank rejects your application. Now, you would like to know why. In particular, you would like to know what would have to be different so that your application would have been accepted.
A possible explanation might be that you would have been accepted if you had earned 500\$ more per month and if you had not had a second credit card.}

\begin{figure}
    \centering
    \includegraphics[scale=.3]{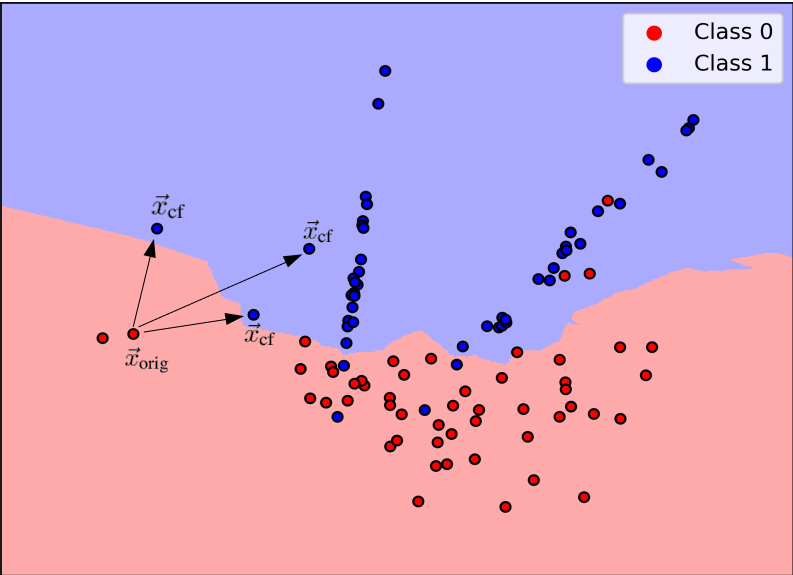}
    \caption{``Rashomon effect'': Illustration of multiple possible counterfactual explanations $\xcf$ of a given sample $\xorig$ for a binary classifier.}
    \label{fig:foundations:counterfactual}
\end{figure}

Unfortunately, many explanation methods (including counterfactual explanations) are lacking uniqueness: Often there exists more than one possible \& valid explanation -- this is called ``Rashomon effect''~\cite{molnar2019} -- and in such cases, it is not clear which or how many of the possible explanations should be presented to the user. See Figure~\ref{fig:foundations:counterfactual} where we illustrate the concept of a counterfactual explanation, including the existing of multiple possible and valid counterfactuals. Most approaches ignore this problem, however, there exist a few approaches that propose to compute multiple diverse counterfactuals to make the user aware that there exist different possible explanations~\cite{rodriguez2021beyond,russell2019efficient,mothilal2020explaining}.
In order to keep the explanation (suggested changes) simple -- i.e. we are looking for low-complexity explanations that are easy to understand -- an obvious strategy is to look for a small number of changes so that the resulting sample (counterfactual) is similar/close to the original sample. This is aimed to be captured by~\refdef{def:counterfactual}.
\begin{definition}[(Closest) Counterfactual Explanation~\cite{CounterfactualWachter}]\label{def:counterfactual}
Assume a prediction function (e.g. a classifier) $\classifier:\RN^\dimsym \to \setY$ is given. Computing a counterfactual $\xcf \in \RN^\dimsym$ for a given input $\x \in \RN^\dimsym$ is phrased as an optimization problem:
\begin{equation}\label{eq:counterfactualoptproblem}
\underset{\xcf \,\in\, \RN^\dimsym}{\arg\min}\; \loss\big(\classifier(\xcf), \ycfclassifier\big) + C \cdot \regularization(\xcf, \x)
\end{equation}
where $\loss(\cdot)$ denotes a loss function, $\ycf$ the target prediction, $\regularization(\cdot)$ a penalty for dissimilarity of $\xcf$ and $\x$, and $C>0$ denotes the regularization strength.
\end{definition}

The counterfactuals from~\refdef{def:counterfactual} are also called \textit{closest counterfactuals} because the optimization problem~\refeq{eq:counterfactualoptproblem} tries to find an explanation $\xcf$ that is as close as possible to the original sample $\x$. However, other aspects like plausibility and actionability are ignored in~\refdef{def:counterfactual}, but are covered in other work~\cite{CounterfactualGuidedByPrototypes,PlausibleCounterfactuals,ActionableCounterfactuals}. In this work, we refer to counterfactuals in the spirit of~\refdef{def:counterfactual}.
Note that counterfactual explanations also exist in the causality domain~\cite{pearl2010causal}. Here the knowledge of a structural causal model (SCM), describing the interaction of features, is assumed. This work is not based in the causality domain and we only consider counterfactual explanations as proposed by~\cite{CounterfactualWachter}.

\section{Counterfactual Explanations of Dimensionality Reduction}\label{sec:cf_dimred}
In this section, we propose counterfactual explanations of dimensionality reduction -- i.e.\ explaining why a specific point was mapped to a location instead of a requested different location. See Fig.\ \ref{fig:idea} for an illustration of this problem. As it is the nature of counterfactual explanations, the explanations state how we have to (minimally) change the original sample such that it gets mapped to a requested location.

\begin{figure*}
    \centering
    \begin{minipage}[b]{0.49\textwidth}
        \includegraphics[width=\textwidth]{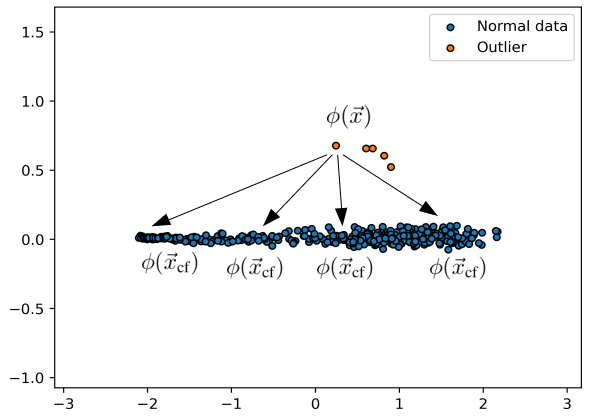}  
   \end{minipage}
     \hfill
    \begin{minipage}[b]{0.49\textwidth}
        \includegraphics[width=\textwidth]{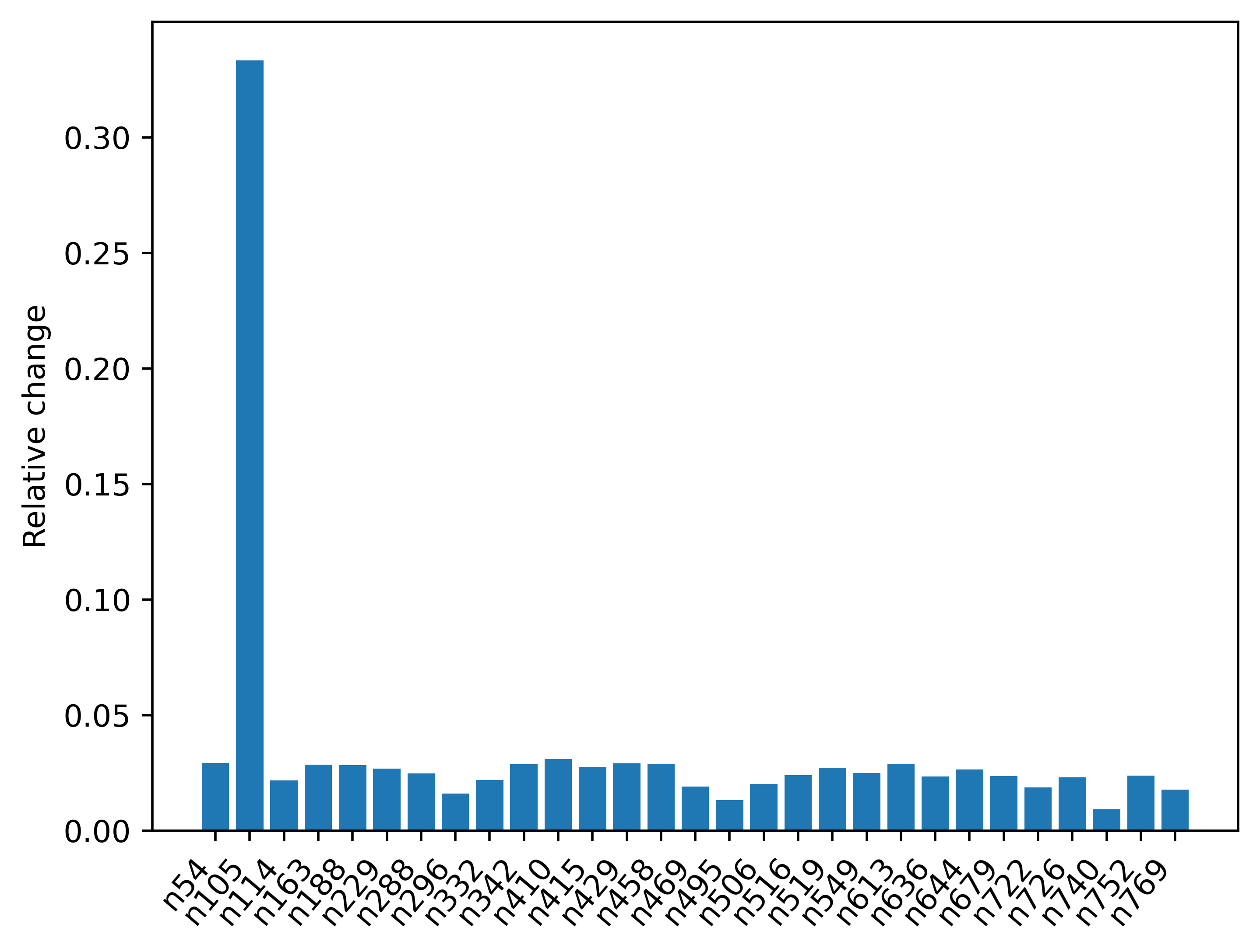}
    \end{minipage}

    \caption{Explaining anomalous pressure measurements -- Left: Two dimensional data visualization of $29$-dimensional pressure measurements; Right: Normalized amount of suggested changes per sensor -- the faulty sensor is suggested to change the most and is therefore correctly identified.}
    \label{fig:cf_dimred:illustration}
\end{figure*}

We argue that this type of explanation is in particular very well suited for explaining data visualization which is a common application of dimensionality reduction in data mining~\cite{gisbrecht2015data,lee2007nonlinear,kaski2011dimensionality} -- e.g. data is mapped to a two-dimensional space which is then depicted in a scatter plot. For instance, we could utilize such explanations to explain outliers in the data visualization: I.e.\ explaining why a point got mapped far away from the other points instead of close to the other ones -- a counterfactual explanation states how to change the outlier such that it is no longer an outlier in the visualization, which would allow us to learn \smth\ about the particular reasons why this point was flagged as an outlier in the visualization.
See Figure~\ref{fig:cf_dimred:illustration} for an illustrative example where we explain anomalous pressure measurements in a water distribution network: We consider the hydraulically isolated ``Area A'' in the L-Town network~\cite{vrachimis2020battledim} where $29$ pressure sensors are installed -- we simulate a sensor failure (constant added the original pressure value) in node \textit{n105}. We pick an outlier $\x$ (see left plot in Figure~\ref{fig:cf_dimred:illustration}) and compute a counterfactual explanation for each normal data point as a target mapping -- i.e. asking which sensor measurements must be changed so that the overall measurement vector $\xcf$ is mapped to the specified location in the data visualization. When aggregating all explanations by summing up and normalizing the suggested changes for each sensor (see right plot in Figure~\ref{fig:cf_dimred:illustration}), we are able to identify the faulty sensors and thereby ``explain'' the outlier.

Note that existing explanation methods for explaining dimensionality reduction methods, which usually focus on feature importances (see Section~\ref{sec:intro}), can not provide such an explanation -- i.e. answering contrasting questions like ``Why was the point mapped here and not there''. This is because they only highlight important features but do not suggest any changes or magnitude of changes that would yield a different (requested) mapping.
However, as already mentioned in Section~\ref{sec:foundations:cf}, the Rashomon effect states that there might exist many possible explanations why a particular point was mapped far away from the others -- we therefore aim for a set of diverse counterfactual explanations in order to learn the most about the observed mapping and provide different possibilities for actionable recourse.

First, we formalize the general concept of (diverse) counterfactual explanations of dimensionality reduction in Section~\ref{sec:modeling}. Next, we consider some popular parametric dimensionality reduction methods, and propose methods for efficiently computing single counterfactuals (see Section~\ref{sec:modelspecific}) and diverse counterfactuals (see Section~\ref{sec:diverse_counterfactuals}).

\subsection{General Modeling}\label{sec:modeling}
We assume the DR method is given as a mapping
\begin{equation}\label{eq:dimred}
    \dimred: \RN^\dimsym \to \RN^{\dimsym'}
\end{equation}
with $\dimsym > \dimsym'$.

A counterfactual explanation of a sample $\x\in\RN^\dimsym$ is a sample $\xcf$ in the original domain (i.e. $\RN^\dimsym$) that differs in a few features only from the given original sample $\x$, but is mapped to a requested location $\ycf\in\RN^{\dimsym'}$ which is different from the mapping of the original sample $\x$.
We formalize this in~\refdef{def:cf_dimred}.
\begin{definition}[Counterfactual Explanation of Dimensionality Reduction]\label{def:cf_dimred}
For a given DR method $\dimred(\cdot)$~\refeq{eq:dimred}, a counterfactual explanation $\xcf\in\RN^\dimsym, \ycf\in\RN^{\dimsym'}$ of a specific sample $\x\in\RN^\dimsym$ is given as a solution to the following multi-criteria optimization problem:
\begin{equation}\label{eq:cf_dimred}
    \underset{\xcf\,\in\,\RN^\dimsym}{\min}\;\Big(\pnorm{\x - \xcf}_0, \pnorm{\dimred(\xcf) - \ycf}_{p}\Big),
\end{equation}
\end{definition}
where $p$ defines the norm that is used.
As discussed in Section~\ref{sec:foundations:cf}, there usually exists more than one possible explanation (``Rashomon effect'') -- clearly this is the case for dimensionality reduction as well because dimensionality reduction is a many-to-one mapping (i.e. multiple points are mapped to the same location). In this context, a set of diverse (i.e. highly different) explanations would provide more information than a single explanation only. We therefore extend~\refdef{def:cf_dimred} to a set of diverse counterfactuals explanations instead of a single one (see~\refdef{def:diverse_cf_dimred}).
\begin{definition}[Diverse Counterfactual Explanations of Dimensionality Reduction]\label{def:diverse_cf_dimred}
For a given DR method $\dimred(\cdot)$~\refeq{eq:dimred}, a set of diverse counterfactual explanations $\{\xcf^i\in\RN^\dimsym\}, \ycf\in\RN^{\dimsym'}$ of a specific sample $\x\in\RN^\dimsym$ is given as a solution to the following multi-criteria optimization problem:
\begin{equation}\label{eq:diverse_cf_dimred}
    \underset{\{{\xcf}^i\,\in\,\RN^\dimsym\}}{\min}\;\Big(\pnorm{\x - \xcf^i}_0, \pnorm{\dimred(\xcf^i) - \ycf}_{p}, \diversity({\xcf}^i, {\xcf}^j)\Big)
\end{equation}
where $\diversity: \RN^\dimsym \times \RN^\dimsym \to \RN_{+}$ denotes a function measuring the pair-wise diversity of two given counterfactuals -- i.e. returning a small value if the two counterfactuals are very different and a larger value otherwise.
\end{definition}
The term ``diversity'' itself is somewhat fuzzy and different use-cases might require different definitions of \textit{diverse counterfactuals}. In this work we utilize a very general definition of diversity, namely the number of overlapping features -- i.e. diverse counterfactuals should not change the same features:
\begin{equation}\label{eq:diversity}
    \diversity({\xcf}^j, {\xcf}^k) = \sum_{i=1}^{\dimsym} \mathds{1}\left(({\deltacf}^j)_i \neq 0 \land ({\deltacf}^k)_i \neq 0\right)
\end{equation}
where ${\deltacf}^j = \xcf^j - \x$ and $\mathds{1}(\cdot)$ denotes the indicator function that returns $1$ if the boolean expression is true and $0$ otherwise.

\subsection{Method Specific Computation of a Single Counterfactual}\label{sec:modelspecific}
Hereinafter, we propose practical relaxations for computing a single counterfactual explanation of different parametric dimensionality reduction methods (see~\refdef{def:cf_dimred}). While~\refdef{def:cf_dimred} does not make any assumptions on the dimensionality reduction $\dimred(\cdot)$, we now assume a parametric dimensionality reduction in order to get tractable optimization problems.

Note that in all cases, we approximate the $0$-norm with the $1$-norm for measuring closeness between the original sample $\x$ and the counterfactual $\xcf$. Furthermore, we use $p=2$, the $2$-norm for measuring the distance between the mapping of $\xcf$ and the requested mapping $\ycf$.

\subsubsection{Linear Methods}
%

In the case of linear mappings as defined in Section~\ref{sec:linMap}, we phrase the computation of a single counterfactual explanations as the following convex quadratic program:
\begin{equation}
\begin{split}
    &\underset{\xcf\,\in\,\RN^\dimsym}{\argmin}\;\pnorm{\x - \xcf}_1 + C \cdot \xi \\
    & \text{s.t. } \pnorm{\mat{A} \xcf + \vec{b} - \ycf}_{2}^2 \leq \xi \\
    & \xi \geq 0
\end{split}
\end{equation}
where $C>0$ acts as a regularization strength balancing between the two objectives in~\refeq{eq:cf_dimred} -- the regularization is necessary because it is numerically difficult (or even impossible) to find a counterfactual $\xcf$ that yields the exact mapping $\dimred(\xcf)=\ycf$, we therefore have to specify how much difference we are willing to tolerate.
Note that convex quadratic programs can be solved efficiently~\cite{Boyd2004}.

\subsubsection{Self Organizing Map}

Similar to linear methods, we phrase the computation of a single counterfactual explanations for SOMs (Section~\ref{sec:SOM}) as the following convex quadratic program, which again can be solved efficiently using standard solvers from convex optimization~\cite{Boyd2004}:
\begin{equation}\label{eq:dimred_cf:som}
\begin{split}
    &\underset{\xcf\,\in\,\RN^\dimsym}{\argmin}\;\pnorm{\x - \xcf}_1 \\
    & \text{s.t. } \pnorm{\xcf - \prototype_{\ycf}}_{2}^2 + \epsilon \leq \pnorm{\xcf - \prototype_{\z}}_{2}^2 \quad \forall\,\z\in\setI
\end{split}
\end{equation}
where $\epsilon > 0$ makes sure that the set of feasible solutions is closed.

\subsubsection{Autoencoder}

For autoencoders (AEs) as discussed in Section~\ref{sec:autoenc}, we utilize the penalty method to merge the two objectives from~\refeq{eq:cf_dimred} into a single objective:
\begin{equation}\label{eq:dimred:ae}
    \underset{\xcf\,\in\,\RN^\dimsym}{\argmin}\;\pnorm{\x - \xcf}_1 + C \cdot \pnorm{\encoder_{\params}(\xcf) - \ycf}_2
\end{equation}
where the hyperparameter $C>0$ acts as a regularization strength.

Assuming continuous differentiability of the encoder $\encoder_{\params}(\cdot)$, we can solve~\refeq{eq:dimred:ae} using a gradient based method. However, due to the non-linearity of $\encoder_{\params}(\cdot)$, we might find a local optimum only.

\subsubsection{Parametric t-SNE}

Although the neural network $\mlp_{\params}(\cdot)$ of parametric t-SNE (Section~\ref{sec:tsne}) is trained in a completely different way compared to an autoencoder based dimensionality reduction, the final modeling is the same and consequently, everything from the case of autoencoder based DR applies here as well:
\begin{equation}
    \underset{\xcf\,\in\,\RN^\dimsym}{\argmin}\;\pnorm{\x - \xcf}_1 + C \cdot \pnorm{\mlp_{\params}(\xcf) - \ycf}_2
\end{equation}

\subsection{Computation of Diverse Counterfactuals}\label{sec:diverse_counterfactuals}
In this section, we propose an algorithm for computing diverse counterfactual explanations (see~\refdef{def:diverse_cf_dimred}) of the four DR methods considered in the previous section. 

Regarding the formalization of diversity~\refeq{eq:diversity}, instead of using~\refeq{eq:diversity} directly, we propose a more stricter version in order to get a continuous function which then yields tractable optimization problems, similar to the ones we proposed in the previous section:
In order to compute a set of diverse counterfactuals instead of a single counterfactual, we utilize our proposed methods for computing a single counterfactual explanations from Section~\ref{sec:modelspecific} and extend these with a mechanism to forbid or punish changes in black-listed features. We then first compute a single counterfactual explanations using the methodology proposed in Section~\ref{sec:modelspecific} and then iteratively compute another counterfactual explanation but black-listing all features that have been changed in the previous counterfactuals -- this procedure is illustrated as pseudo-code in Algorithm~\ref{algo:diverse_cf}.
\begin{algorithm}[t]
\caption{Computation of Diverse Counterfactuals}\label{algo:diverse_cf}
\textbf{Input:} Original input $\x$, Target location $\ycf$, $k\geq 1$: number of diverse counterfactuals, Dimensionality reduction $\dimred(\cdot)$ \\
\textbf{Output:} Set of diverse counterfactuals $\set{R}=\{\xcf^i\}$
\begin{algorithmic}[1]
 \State $\setF = \{\}$  \Comment{Initialize set of black-listed features}
 \State $\set{R} = \{\}$  \Comment{Initialize set of diverse counterfactuals}
 \For{$i=1,\dots,k$} \Comment{Compute $k$ diverse counterfactuals}
    \State $\xcf^i = \myCF{\x}{\ycf}{\set{F}}{\dimred}$  \Comment{Compute next counterfactual}
    \State $\set{R} = \set{R} \cup \{\xcf^i\}$
    \State $\setF = \setF \cup \{j \mid (\xcf - \x)_j \neq 0\}$  \Comment{Update set of black-listed features}
 \EndFor
\end{algorithmic}
\end{algorithm}
\paragraph*{Black-listing features}
We assume we are given an ordered set $\setF$ of black-listed features.
In case of convex programs (e.g. linear methods and SOM), we consider black-listed features $\setF$ by means of an additional affine equality constraint:
\begin{equation}
    \mat{M}\xcf = \vec{m}
\end{equation}
where $\mat{M}\in\RN^{|\set{F}| \times \dimsym}, \vec{m}\in\RN^{|\set{F}|}$ with
\begin{equation}
    (\mat{M})_{i,j} =
    \begin{cases}
    1  & \quad \text{if } (\setF)_i = j\\
    0  & \quad \text{otherwise}
  \end{cases}
\end{equation}
and $\vec{m}_k = (\x)_{(\setF)_k}$.

Whereas in all other cases (e.g. autoencoder and parametric t-SNE), where we minimize a (non-convex) cost function, we replace the counterfactual $\xcf$ in the optimization problem with an affine mapping undoing any potential changes in black-listed features -- i.e. black-listed features can be changed but have no effect on the final counterfactual because they are reset to their original value:
\begin{equation}
    \pnorm{\dimred(\mat{M} \xcf + \vec{m}) - \ycf}_2
\end{equation}
where $\mat{M}\in\RN^{\dimsym\times\dimsym}, \vec{m}\in\RN^\dimsym$ with
\begin{equation}
\begin{split}
    &(\mat{M})_{i,j} =
    \begin{cases}
    1  & \quad \text{if } i=j \text{ and } i \not\in \setF\\
    0  & \quad \text{otherwise}
  \end{cases}
  \\
  &(\vec{m})_{i} =
    \begin{cases}
    (\x)_i  & \quad \text{if } i\in\setF\\
    0  & \quad \text{otherwise}
  \end{cases}
\end{split}
\end{equation}
Note that in both cases, the complexity and type of optimization problem does not change -- e.g. convex programs remain convex programs.

For convenience, we use $\myCF{\x}{\ycf}{\set{F}}{\dimred}$ to denote the computation of a counterfactual $(\xcf,\ycf)$ of a DR method $\dimred(\cdot)$ at a given sample $\x$ subject to a set $\set{F}$ of black-listed features.

\section{Experiments}\label{sec:experiments}
We empirically evaluate our proposed explanation methodology of DR methods on the specific use-case of data visualization -- i.e. dimensionality reduction to two dimensions. All experiments are implemented in Python and are publicly available on GitHub\footnote{\url{https://github.com/andreArtelt/ContrastingExplanationDimRed}}.

\subsection{Data}
We run all our experiments on a set of different ML benchmark data sets -- all data sets are standardized:
\paragraph{Diabetes} The ``Diabetes Data Set''~\cite{diabetes} is a labeled data set containing recordings from diabetes patients. The data set contains $442$ samples and $10$ real valued scaled features in $[-.2, .2]$ such as body mass index, age in years and average blood pressure. The labels are integers in $[25, 346]$ denoting a quantitative measure of disease progression one year after baseline.
\paragraph{Breast cancer} The ``Breast Cancer Wisconsin (Diagnostic) Data Set''~\cite{breastcancer} is used for classifying breast cancer samples into benign and malignant (i.e. binary classification). The data set contains $569$ samples and $30$ numerical features such as area, smoothness and compactness.
\paragraph{Toy} An an artificial, self created, toy data set containing $500$ ten dimensional samples. Each feature is distributed according to a normal distribution whereby we choose a different random mean for each feature - by this we can guarantee that, in contrast to the other data sets, the features are independent of each other. The binary labelling of the samples is done by splitting the data into two clusters using k-means.

\subsection{Model Agnostic Algorithm for Comparison}\label{sec:exp:baseline}
We compare Algorithm~\ref{algo:diverse_cf} to a general model agnostic algorithm (ModelAgnos) for computing diverse counterfactual explanations where we select samples from the training data set $\set{D}$ that minimize a weighted combination of~\refeq{eq:diverse_cf_dimred} -- i.e. we make use of the penalty method to solve the multi-objective optimization problem~\refeq{eq:diverse_cf_dimred} without making any further assumption on the dimensionality reduction $\dimred(\cdot)$:
\begin{equation}
        \underset{\{{\xcf}^i\,\in\,\set{D}\}}{\min}\;C_1 \cdot \pnorm{\x - \xcf^i}_1 + C_2 \cdot \pnorm{\dimred(\xcf^i) - \ycf}_{2} + C_3 \cdot \diversity({\xcf}^i, {\xcf}^j)
\end{equation}
where $C_1, C_2, C_3 > 0$ denote regularization coefficients that allow us to balance between the different objectives, and $\diversity(\cdot)$ is implemented as stated in~\refeq{eq:diversity}.
By limiting the set of feasible solutions to the training data set, we can guarantee plausibility of the resulting counterfactual explanations -- note that plausibility of the counterfactuals generated by Algorithm~\ref{algo:diverse_cf} can not be guaranteed.

\subsection{Setup}\label{sec:exp:setup}
For each data set and each of the four parametric DR methods (PCA, Autoencoder, SOM, parametric t-SNE) from Section~\ref{sec:modelspecific}, we fit the DR method to the entire data set and compute for each sample in the data set a set of three diverse counterfactual explanations -- we evaluate and compare the counterfactuals\footnote{All hyperparameters (regularization strength) $C_{i}\forall i$ are set to $1$.} computed by our proposed Algorithm~\ref{algo:diverse_cf} with those from the model agnostic algorithm (see Section~\ref{sec:exp:baseline}). For the requested target location $\ycf$ -- recall that in a counterfactual explanation we ask for a change that would lead to a different specified mapping $\ycf$ instead of the original mapping $\y$ -- we either choose the mapping of a different sample (with a different label) from the training data set as $\ycf$ or the mapping of the original sample $\x$ after perturbing three random features -- the same type of perturbation is applied to these three features. Regarding the perturbations, we consider the following ones:
\begin{itemize}
    \item \textit{Shift}: A constant is added to the feature value.
    \item \textit{Gaussian}: Gaussian noise is added to feature value.
\end{itemize}
Note that we evaluate each perturbation separately. Furthermore, note that these perturbations could be interpreted as sensor failures and are therefore highly relevant to practice.

\subsection{Evaluation}\label{sec:exp:metrics}
For all experimental scenarios, we monitor and evaluate some quantitative measurements:
\begin{itemize}\label{itemize:exp_eval_metrics}
    \item \textit{CfSparse}: Sparsity of the counterfactual explanations -- i.e. how many (percentage) of the available features are used in the explanation, smaller values are better.
    \item \textit{CfDist}: Euclidean distance between the mapping of the counterfactual $\dimred(\xcf)$ and the requested mapping $\ycf$ -- i.e. this can be interpreted as a measurement of the error of counterfactual explanations, smaller values are better.
    \item \textit{CfDiv}: Diversity of the counterfactual explanations -- i.e. the number of overlapping features between the diverse explanations (see~\refeq{eq:diversity} in Section~\ref{sec:modeling}), smaller values are better.
\end{itemize}
For the scenarios where we apply a perturbation to the original sample, we also record the recall of the identified perturbed features in the counterfactual explanations -- i.e. checking if the used features in the explanation coincide with the perturbed features. By this, we try to measure the usefulness of our explanations for identifying relevant features -- however, since dimensionality reduction is a many-to-one mapping, we consider recall only because we do not expect to observe a high precision due to the Rashomon effect.

Note that each experiment is repeated $100$ times in order to get statistically reliable estimates of the quantitative measurements.

\subsection{Results}\label{sec:exp:results}
\begin{table*}
\caption{Quantitative results: \textit{No perturbation} -- all numbers are rounded to two decimal places, best scores are highlighted in \textbf{bold-face}.}
\centering
\footnotesize
\begin{tabular}{|c|c||c|c||c|c||c|c||}
 \hline
 & \multirow{2}{*}{\textit{DataSet}} & \multicolumn{2}{c||}{CfSparse $\downarrow$} & \multicolumn{2}{c||}{CfDiv $\downarrow$} & \multicolumn{2}{c||}{CfDist $\downarrow$} \\
 &  & Algo~\ref{algo:diverse_cf} & ModelAgnos & Algo~\ref{algo:diverse_cf} & ModelAgnos & Algo~\ref{algo:diverse_cf} & ModelAgnos \\
 \hline
 \multirow{3}{*}{\rotatebox[origin=c]{90}{Linear}}
& Diabetes & $\mathbf{0.21 \pm 0.0}$ & $0.55 \pm 0.0$ & $\mathbf{0.0 \pm 0.0}$ & $7.07 \pm 0.86$ & $\mathbf{0.2 \pm 0.2}$ & $1.33 \pm 0.39$\\
& Breast cancer & $\mathbf{0.15 \pm 0.01}$ & $0.66 \pm 0.0$ & $\mathbf{0.0 \pm 0.0}$ & $29.6 \pm 0.89$ & $\mathbf{0.36 \pm 1.47}$ & $2.64 \pm 1.5$\\
& Toy & $\mathbf{0.21 \pm 0.0}$ & $0.67 \pm 0.0$ & $\mathbf{0.0 \pm 0.0}$ & $9.99 \pm 0.01$ & $\mathbf{0.03 \pm 0.04}$ & $0.85 \pm 0.12$ \\
 \hline\hline
 \multirow{3}{*}{\rotatebox[origin=c]{90}{SOM}}
& Diabetes & $0.88 \pm 0.02$ & $\mathbf{0.61 \pm 0.01}$ & $\mathbf{0.0 \pm 0.0}$ & $9.63 \pm 21.16$ & $\mathbf{0.01 \pm 0.15}$ & $3.2 \pm 2.73$\\
& Breast cancer & $0.91 \pm 0.02$ & $\mathbf{0.66 \pm 0.0}$ & $\mathbf{0.0 \pm 0.04}$ & $29.71 \pm 0.62$ & $\mathbf{0.4 \pm 5.22}$ & $4.07 \pm 3.91$\\
& Toy & $0.84 \pm 0.03$ & $\mathbf{0.68 \pm 0.0}$ & $\mathbf{0.0 \pm 0.0}$ & $10.6 \pm 11.73$ & $\mathbf{0.01 \pm 0.17}$ & $3.77 \pm 3.18$\\
 \hline\hline
 \multirow{3}{*}{\rotatebox[origin=c]{90}{AE}}
& Diabetes & $\mathbf{0.14 \pm 0.03}$ & $0.51 \pm 0.0$ & $\mathbf{0.0 \pm 0.01}$ & $6.15 \pm 0.7$ & $0.28 \pm 0.08$ & $\mathbf{0.23 \pm 0.04}$\\
& Breast cancer & $\mathbf{0.03 \pm 0.01}$ & $0.65 \pm 0.0$ & $\mathbf{0.0 \pm 0.05}$ & $28.95 \pm 0.92$ & $0.36 \pm 0.16$ & $\mathbf{0.31 \pm 0.12}$\\
& Toy & $\mathbf{0.14 \pm 0.03}$ & $0.67 \pm 0.0$ & $\mathbf{0.0 \pm 0.02}$ & $9.98 \pm 0.02$ & $0.3 \pm 0.09$ & $\mathbf{0.18 \pm 0.02}$\\
 \hline\hline
 \multirow{3}{*}{\rotatebox[origin=c]{90}{t-SNE}}
& Diabetes & $\mathbf{0.33 \pm 0.0}$ & $0.58 \pm 0.0$ & $\mathbf{0.0 \pm 0.0}$ & $8.12 \pm 1.11$ & $5.35 \pm 7.43$ & $\mathbf{3.0 \pm 1.83}$\\
& Breast cancer & $\mathbf{0.32 \pm 0.01}$ & $0.67 \pm 0.0$ & $\mathbf{0.23 \pm 4.74}$ & $29.86 \pm 0.17$ & $8.52 \pm 11.59$ & $\mathbf{4.32 \pm 2.16}$\\
& Toy & $\mathbf{0.33 \pm 0.0}$ & $0.67 \pm 0.0$ & $\mathbf{0.0 \pm 0.0}$ & $10.0 \pm 0.0$ & $1.92 \pm 0.81$ & $\mathbf{1.06 \pm 0.21}$ \\
 \hline
\end{tabular}
\label{table:experimentresults:perturbation:non}
\end{table*}

\begin{table*}
\caption{Quantitative results: \textit{Shift perturbation} -- all numbers are rounded to two decimal places, best scores are highlighted in \textbf{bold-face}.}
\centering
\footnotesize
\begin{tabular}{|c|c||c|c||c|c||c|c||c|c||}
 \hline
 & \multirow{2}{*}{\textit{DataSet}} & \multicolumn{2}{c||}{CfSparse $\downarrow$} & \multicolumn{2}{c||}{CfDiv $\downarrow$} & \multicolumn{2}{c||}{CfDist $\downarrow$}  & \multicolumn{2}{c||}{Recall $\uparrow$} \\
 &  & Algo~\ref{algo:diverse_cf} & ModelAgnos & Algo~\ref{algo:diverse_cf} & ModelAgnos & Algo~\ref{algo:diverse_cf} & ModelAgnos & Algo~\ref{algo:diverse_cf} & ModelAgnos \\
 \hline
 \multirow{3}{*}{\rotatebox[origin=c]{90}{Linear}}
& Diabetes & $\mathbf{0.23 \pm 0.0}$ & $0.56 \pm 0.0$ & $\mathbf{0.0 \pm 0.0}$ & $7.44 \pm 0.82$ & $\mathbf{0.28 \pm 0.37}$ & $2.86 \pm 1.42$ & $0.72 \pm 0.07$ & $\mathbf{0.95 \pm 0.02}$\\
& Breast cancer & $\mathbf{0.11 \pm 0.01}$ & $0.66 \pm 0.0$ & $\mathbf{0.0 \pm 0.0}$ & $29.48 \pm 1.02$ & $\mathbf{0.02 \pm 0.02}$ & $1.56 \pm 0.26$ & $0.42 \pm 0.09$ & $\mathbf{1.0 \pm 0.0}$\\
& Toy & $\mathbf{0.21 \pm 0.0}$ & $0.67 \pm 0.0$ & $\mathbf{0.0 \pm 0.0}$ & $10.0 \pm 0.0$ & $\mathbf{0.01 \pm 0.03}$ & $2.34 \pm 2.21$ & $0.8 \pm 0.04$ & $\mathbf{1.0 \pm 0.0}$\\
 \hline\hline
 \multirow{3}{*}{\rotatebox[origin=c]{90}{SOM}}
& Diabetes & $0.8 \pm 0.04$ & $\mathbf{0.6 \pm 0.01}$ & $\mathbf{0.0 \pm 0.01}$ & $9.31 \pm 20.54$ & $\mathbf{0.01 \pm 0.09}$ & $3.74 \pm 3.72$ & $0.94 \pm 0.02$ & $\mathbf{0.96 \pm 0.01}$\\
& Breast cancer & $0.67 \pm 0.07$ & $\mathbf{0.66 \pm 0.0}$ & $\mathbf{0.01 \pm 0.18}$ & $29.68 \pm 0.84$ & $\mathbf{0.31 \pm 4.04}$ & $4.6 \pm 13.18$ & $0.95 \pm 0.02$ & $\mathbf{1.0 \pm 0.0}$\\
& Toy & $0.77 \pm 0.04$ & $\mathbf{0.67 \pm 0.0}$ & $\mathbf{0.0 \pm 0.01}$ & $10.47 \pm 9.14$ & $\mathbf{0.02 \pm 0.25}$ & $3.67 \pm 3.3$ & $0.92 \pm 0.03$ & $\mathbf{1.0 \pm 0.0}$\\
 \hline\hline
 \multirow{3}{*}{\rotatebox[origin=c]{90}{AE}}
& Diabetes & $\mathbf{0.14 \pm 0.03}$ & $0.51 \pm 0.0$ & $\mathbf{0.0 \pm 0.03}$ & $6.21 \pm 0.71$ & $0.78 \pm 0.6$ & $\mathbf{0.74 \pm 0.63}$ & $0.43 \pm 0.24$ & $\mathbf{0.9 \pm 0.03}$\\
& Breast cancer & $\mathbf{0.04 \pm 0.01}$ & $0.65 \pm 0.0$ & $\mathbf{0.0 \pm 0.04}$ & $28.98 \pm 0.94$ & $\mathbf{0.47 \pm 0.24}$ & $0.51 \pm 0.24$ & $0.13 \pm 0.1$ & $\mathbf{1.0 \pm 0.0}$\\
& Toy & $\mathbf{0.16 \pm 0.03}$ & $0.67 \pm 0.0$ & $\mathbf{0.0 \pm 0.01}$ & $9.97 \pm 0.03$ & $0.62 \pm 0.26$ & $\mathbf{0.57 \pm 0.39}$ & $0.48 \pm 0.25$ & $\mathbf{1.0 \pm 0.0}$\\
 \hline\hline
 \multirow{3}{*}{\rotatebox[origin=c]{90}{t-SNE}}
& Diabetes & $\mathbf{0.33 \pm 0.0}$ & $0.58 \pm 0.0$ & $\mathbf{0.01 \pm 0.09}$ & $8.06 \pm 0.91$ & $6.13 \pm 8.14$ & $\mathbf{4.4 \pm 5.63}$ & $\mathbf{1.0 \pm 0.0}$ & $0.96 \pm 0.01$\\
& Breast cancer & $\mathbf{0.32 \pm 0.0}$ & $0.66 \pm 0.0$ & $\mathbf{0.08 \pm 1.42}$ & $29.65 \pm 0.95$ & $3.14 \pm 2.2$ & $\mathbf{2.0 \pm 0.55}$ & $0.97 \pm 0.02$ & $\mathbf{1.0 \pm 0.0}$\\
& Toy & $\mathbf{0.33 \pm 0.0}$ & $0.67 \pm 0.0$ & $\mathbf{0.0 \pm 0.0}$ & $10.0 \pm 0.0$ & $2.81 \pm 1.46$ & $\mathbf{1.87 \pm 0.97}$ & $\mathbf{1.0 \pm 0.0}$ & $\mathbf{1.0 \pm 0.0}$\\
 \hline
\end{tabular}
\label{table:experimentresults:perturbation:shift}
\end{table*}
The results of the scenario without any perturbations -- i.e. randomly selecting the target sample from the training set -- are shown in Table~\ref{table:experimentresults:perturbation:non} and the results of the scenarios with perturbations are shown in Tables~\ref{table:experimentresults:perturbation:gaussian},\ref{table:experimentresults:perturbation:shift} -- note that, due to space constraints, the latter one is put in the appendix.

We observe that Algorithm~\ref{algo:diverse_cf}, on average, achieves much sparser and more diverse explanations than the mode agnostic algorithm (Section~\ref{sec:exp:baseline}) does. Only in case of SOM, the sparsity is often a bit worse than those from the baseline -- this might be due to numerical instabilities of the mathematical program~\refeq{eq:dimred_cf:som}. In particular, while Algorithm~\ref{algo:diverse_cf} almost always yields completely diverse explanations, the model agnostic algorithm fails completely -- this highlights the strength of our proposed Algorithm~\ref{algo:diverse_cf} for computing diverse explanations.
Furthermore, both methods are able to yield counterfactual explanations that are very close to the requested target location. In most cases Algorithm~\ref{algo:diverse_cf} yields counterfactuals that are closer to the target location, only in case of parametric t-SNE the model agnostic algorithm yields ``better'' counterfactuals -- however, in both cases the variance is quite large which indicates instabilities of the learned dimensionality reduction.
Note that, since the three evaluation metrics~\ref{itemize:exp_eval_metrics} are contradictory, it can be misleading to evaluate the performance under each metric separately without looking at the other metrics at the same time -- e.g. a method might yield very sparse counterfactuals but their distance to the requesting mappings is very large. In order to compensate the contradictory nature of the evaluation metrics, we suggest to also consider a ranking over the three metrics when assessing the performance of the two proposed algorithms for computing counterfactuals -- we give such a ranking in Tables~\ref{table:experimentresults:ranking:perturbation:non},\ref{table:experimentresults:ranking:perturbation:shift},\ref{table:experimentresults:ranking:perturbation:gaussian}. According to these rankings, Algorithm~\ref{algo:diverse_cf} outperforms the model agnostic algorithm in many cases or is at at least as good as the model agnostic method but never worse.
\begin{table}
\caption{Ranking of results from Table~\ref{table:experimentresults:perturbation:non} -- counting the number of metrics where the method yields the best score, best scores are highlighted in \textbf{bold-face}.}
\centering
\footnotesize
\begin{tabular}{|c|c||c|c||}
 \hline
 & \textit{DataSet} & Algo~\ref{algo:diverse_cf} & ModelAgnos \\
 \hline
 \multirow{3}{*}{\rotatebox[origin=c]{90}{Linear}}
    & Diabetes & $\mathbf{3/3}$ & $0/3$ \\
    & Breast cancer & $\mathbf{3/3}$ & $0/3$ \\
    & Toy & $\mathbf{3/3}$ & $0/3$ \\
 \hline\hline
 \multirow{3}{*}{\rotatebox[origin=c]{90}{SOM}}
    & Diabetes & $\mathbf{2/3}$ & $1/3$ \\
    & Breast cancer & $\mathbf{2/3}$ & $1/3$ \\
    & Toy & $\mathbf{2/3}$ & $1/3$ \\
 \hline\hline
 \multirow{3}{*}{\rotatebox[origin=c]{90}{AE}}
    & Diabetes & $\mathbf{2/3}$ & $1/3$ \\
    & Breast cancer & $\mathbf{2/3}$ & $1/3$ \\
    & Toy & $\mathbf{2/3}$ & $1/3$ \\
 \hline\hline
 \multirow{3}{*}{\rotatebox[origin=c]{90}{t-SNE}}
    & Diabetes & $\mathbf{2/3}$ & $1/3$ \\
    & Breast cancer & $\mathbf{2/3}$ & $1/3$ \\
    & Toy & $\mathbf{2/3}$ & $1/3$ \\
 \hline
\end{tabular}
\label{table:experimentresults:ranking:perturbation:non}
\end{table}
\begin{table}
\caption{Ranking of results from Table~\ref{table:experimentresults:perturbation:shift} -- counting the number of metrics where the method yields the best score, best scores are highlighted in \textbf{bold-face}.}
\centering
\footnotesize
\begin{tabular}{|c|c||c|c||}
 \hline
 & \textit{DataSet} & Algo~\ref{algo:diverse_cf} & ModelAgnos \\
 \hline
 \multirow{3}{*}{\rotatebox[origin=c]{90}{Linear}}
    & Diabetes & $\mathbf{3/4}$ & $1/4$ \\
    & Breast cancer & $\mathbf{3/4}$ & $1/4$ \\
    & Toy & $\mathbf{3/4}$ & $1/4$ \\
 \hline\hline
 \multirow{3}{*}{\rotatebox[origin=c]{90}{SOM}}
    & Diabetes & $\mathbf{2/4}$ & $\mathbf{2/4}$ \\
    & Breast cancer & $\mathbf{2/4}$ & $\mathbf{2/4}$ \\
    & Toy & $\mathbf{2/4}$ & $\mathbf{2/4}$ \\
 \hline\hline
 \multirow{3}{*}{\rotatebox[origin=c]{90}{AE}}
    & Diabetes & $\mathbf{2/4}$ & $\mathbf{2/4}$ \\
    & Breast cancer & $\mathbf{3/4}$ & $1/4$ \\
    & Toy & $\mathbf{2/4}$ & $\mathbf{2/4}$ \\
 \hline\hline
 \multirow{3}{*}{\rotatebox[origin=c]{90}{t-SNE}}
    & Diabetes & $\mathbf{3/4}$ & $1/4$ \\
    & Breast cancer & $\mathbf{2/4}$ & $\mathbf{2/4}$ \\
    & Toy & $\mathbf{3/4}$ & $1/4$ \\
 \hline
\end{tabular}
\label{table:experimentresults:ranking:perturbation:shift}
\end{table}
While the recall of the baseline is very good across all DR methods and data sets, the recall of Algorithm~\ref{algo:diverse_cf} is often very good as well, however, there exist some cases (in particular the breast cancer data set) where the recall drops significantly compared to the model agnostic algorithm.

\section{Conclusion}\label{sec:conclusion}
In this work, we proposed the abstract concept of contrasting explanations for locally explaining dimensionality reduction methods -- we considered two-dimensional data visualization as a popular example application. In order to deal with the Rashomon effect -- i.e. the fact that there exist more than one possible and valid explanation -- we considered a set of diverse explanations instead of a single explanation.
Furthermore, we also proposed an implementation of this concept using counterfactual explanations and proposed modelings and algorithms for efficiently computing diverse counterfactual explanations of different parametric dimensionality reduction methods.
We empirically evaluated different aspects of our proposed algorithms on different standard benchmark data sets -- we observe that our proposed methods consistently yield good results.

Based on this initial work, there are a couple of potential extensions and directions for future research:

Depending on the domain and application, it might be necessary to guarantee plausibility of the counterfactuals -- i.e. making sure that the counterfactual $\xcf$ is reasonable and plausible in the data domain. Implausibility or a lack of realism of the counterfactual $\xcf$ might hinder successful recourse in practice. In this work, we ignored the aspect of plausibility and it might happen that the computed counterfactuals $\xcf$ are not always realistic samples from the data domain -- only in case of our model agnostic algorithm (see Section~\ref{sec:exp:baseline}) we can guarantee plausibility because we only consider samples from the training data set $\set{D}$ as potential counterfactuals $\xcf$. In future work, a first approach could be to add plausibility constraints to our proposed modelings (see Section~\ref{sec:modeling}) like it was done for counterfactual explanations of classifiers~\cite{PlausibleCounterfactuals,ActionableCounterfactuals,CounterfactualGuidedByPrototypes}.

Another crucial aspects of transparency \& explainability is the human. In particular, quantitative evaluation of algorithmic properties do not necessary coincide with a human evaluation~\cite{kuhl2022keep}. Therefore we suggest to conduct a user-study to evaluate how ``useful'' our proposed explanation actually are -- in particular it would be of interest to compare normal vs. plausible explanations, and to compare diverse explanations vs. a single explanations.


\bibliographystyle{IEEEtran}
\bibliography{bibliography.bib}

\appendix
\section{Results of the empirical evaluation}
\begin{table*}
\caption{Quantitative results: \textit{Gaussian perturbation} -- all numbers are rounded to two decimal places, best scores are highlighted in \textbf{bold-face}.}
\centering
\footnotesize
\begin{tabular}{|c|c||c|c||c|c||c|c||c|c||}
 \hline
 & \multirow{2}{*}{\textit{DataSet}} & \multicolumn{2}{c||}{CfSparse $\downarrow$} & \multicolumn{2}{c||}{CfDiv $\downarrow$} & \multicolumn{2}{c||}{CfDist $\downarrow$}  & \multicolumn{2}{c||}{Recall $\uparrow$} \\
 &  & Algo~\ref{algo:diverse_cf} & ModelAgnos & Algo~\ref{algo:diverse_cf} & ModelAgnos & Algo~\ref{algo:diverse_cf} & ModelAgnos & Algo~\ref{algo:diverse_cf} & ModelAgnos \\
 \hline
 \multirow{3}{*}{\rotatebox[origin=c]{90}{Linear}}
& Diabetes & $\mathbf{0.22 \pm 0.0}$ & $0.55 \pm 0.0$ & $\mathbf{0.0 \pm 0.0}$ & $7.14 \pm 0.93$ & $\mathbf{0.57 \pm 2.23}$ & $4.24 \pm 49.13$ & $0.7 \pm 0.07$ & $\mathbf{0.93 \pm 0.02}$\\
& Breast cancer & $\mathbf{0.09 \pm 0.0}$ & $0.66 \pm 0.0$ & $\mathbf{0.0 \pm 0.0}$ & $29.43 \pm 1.03$ & $\mathbf{0.05 \pm 0.21}$ & $1.35 \pm 1.7$ & $0.36 \pm 0.09$ & $\mathbf{1.0 \pm 0.0}$\\
& Toy & $\mathbf{0.22 \pm 0.0}$ & $0.67 \pm 0.0$ & $\mathbf{0.0 \pm 0.0}$ & $10.0 \pm 0.0$ & $\mathbf{0.16 \pm 0.62}$ & $2.99 \pm 12.53$ & $0.57 \pm 0.12$ & $\mathbf{1.0 \pm 0.0}$\\
 \hline\hline
 \multirow{3}{*}{\rotatebox[origin=c]{90}{SOM}}
& Diabetes & $0.79 \pm 0.05$ & $\mathbf{0.6 \pm 0.01}$ & $\mathbf{0.0 \pm 0.01}$ & $9.15 \pm 19.24$ & $\mathbf{0.02 \pm 0.2}$ & $3.84 \pm 3.93$ & $0.92 \pm 0.03$ & $\mathbf{0.94 \pm 0.02}$\\
& Breast cancer & $0.6 \pm 0.07$ & $\mathbf{0.66 \pm 0.0}$ & $\mathbf{0.01 \pm 0.2}$ & $29.66 \pm 0.9$ & $\mathbf{0.28 \pm 3.74}$ & $3.87 \pm 14.26$ & $0.95 \pm 0.02$ & $\mathbf{1.0 \pm 0.0}$\\
& Toy & $0.74 \pm 0.06$ & $\mathbf{0.68 \pm 0.0}$ & $\mathbf{0.0 \pm 0.01}$ & $10.55 \pm 10.66$ & $\mathbf{0.01 \pm 0.09}$ & $3.68 \pm 3.82$ & $0.9 \pm 0.04$ & $\mathbf{1.0 \pm 0.0}$\\
 \hline\hline
 \multirow{3}{*}{\rotatebox[origin=c]{90}{AE}}
& Diabetes & $\mathbf{0.14 \pm 0.03}$ & $0.51 \pm 0.0$ & $\mathbf{0.0 \pm 0.01}$ & $6.23 \pm 0.69$ & $1.09 \pm 5.81$ & $\mathbf{1.07 \pm 3.78}$ & $0.43 \pm 0.24$ & $\mathbf{0.91 \pm 0.03}$\\
& Breast cancer & $\mathbf{0.04 \pm 0.01}$ & $0.65 \pm 0.0$ & $\mathbf{0.0 \pm 0.03}$ & $28.98 \pm 0.94$ & $0.48 \pm 0.51$ & $\mathbf{0.38 \pm 0.22}$ & $0.11 \pm 0.09$ & $\mathbf{1.0 \pm 0.0}$\\
& Toy & $\mathbf{0.14 \pm 0.03}$ & $0.67 \pm 0.0$ & $\mathbf{0.0 \pm 0.01}$ & $9.98 \pm 0.02$ & $0.99 \pm 3.21$ & $\mathbf{0.5 \pm 0.99}$ & $0.42 \pm 0.24$ & $\mathbf{1.0 \pm 0.0}$\\
 \hline\hline
 \multirow{3}{*}{\rotatebox[origin=c]{90}{t-SNE}}
& Diabetes & $\mathbf{0.33 \pm 0.0}$ & $0.56 \pm 0.0$ & $\mathbf{0.01 \pm 0.07}$ & $7.78 \pm 1.11$ & $4.69 \pm 11.37$ & $\mathbf{3.78 \pm 8.53}$ & $\mathbf{1.0 \pm 0.0}$ & $0.93 \pm 0.02$\\
& Breast cancer & $\mathbf{0.32 \pm 0.0}$ & $0.66 \pm 0.0$ & $\mathbf{0.05 \pm 0.78}$ & $29.6 \pm 0.97$ & $1.94 \pm 1.99$ & $\mathbf{1.41 \pm 0.64}$ & $0.96 \pm 0.04$ & $\mathbf{1.0 \pm 0.0}$\\
& Toy & $\mathbf{0.33 \pm 0.0}$ & $0.67 \pm 0.0$ & $\mathbf{0.0 \pm 0.02}$ & $10.0 \pm 0.0$ & $2.16 \pm 1.37$ & $\mathbf{1.52 \pm 0.85}$ & $\mathbf{1.0 \pm 0.0}$ & $\mathbf{1.0 \pm 0.0}$\\
 \hline
\end{tabular}
\label{table:experimentresults:perturbation:gaussian}
\end{table*}
\begin{table}
\caption{Ranking of results from Table~\ref{table:experimentresults:perturbation:gaussian} -- counting the number of metrics where the method yields the best score, best scores are highlighted in \textbf{bold-face}.}
\centering
\footnotesize
\begin{tabular}{|c|c||c|c||}
 \hline
 & \textit{DataSet} & Algo~\ref{algo:diverse_cf} & ModelAgnos \\
 \hline
 \multirow{3}{*}{\rotatebox[origin=c]{90}{Linear}}
    & Diabetes & $\mathbf{3/4}$ & $1/4$ \\
    & Breast cancer & $\mathbf{3/4}$ & $1/4$ \\
    & Toy & $\mathbf{3/4}$ & $1/4$ \\
 \hline\hline
 \multirow{3}{*}{\rotatebox[origin=c]{90}{SOM}}
    & Diabetes & $\mathbf{2/4}$ & $\mathbf{2/4}$ \\
    & Breast cancer & $\mathbf{2/4}$ & $\mathbf{2/4}$ \\
    & Toy & $\mathbf{2/4}$ & $\mathbf{2/4}$ \\
 \hline\hline
 \multirow{3}{*}{\rotatebox[origin=c]{90}{AE}}
    & Diabetes & $\mathbf{2/4}$ & $\mathbf{2/4}$ \\
    & Breast cancer & $\mathbf{2/4}$ & $\mathbf{2/4}$ \\
    & Toy & $\mathbf{2/4}$ & $\mathbf{2/4}$ \\
 \hline\hline
 \multirow{3}{*}{\rotatebox[origin=c]{90}{t-SNE}}
    & Diabetes & $\mathbf{3/4}$ & $1/4$ \\
    & Breast cancer & $\mathbf{2/4}$ & $\mathbf{2/4}$ \\
    & Toy & $\mathbf{3/4}$ & $1/4$ \\
 \hline
\end{tabular}
\label{table:experimentresults:ranking:perturbation:gaussian}
\end{table}

\end{document}